%% file: main.tex
\pdfoutput=1
\documentclass[10pt,twocolumn,letterpaper]{article}
\PassOptionsToPackage{table}{xcolor}

\usepackage[pagenumbers]{cvpr}
\usepackage{xspace}

\input{preamble}

\definecolor{cvprblue}{rgb}{0.21,0.49,0.74}
\usepackage[breaklinks,colorlinks,allcolors=cvprblue]{hyperref}

\title{SGAR: Feed-Forward 3D Reconstruction via Sparse Geometry-Guided Attention}

\author{Aoxiang Fan$^{*}$ \hspace{0.5em} Maria Pilligua$^{*}$ \hspace{0.5em} Corentin Dumery \hspace{0.5em} Oliver Heinimann \hspace{0.5em} Pascal Fua\\[0.4em]
CVLab, EPFL, Switzerland\\
{\tt\small \{aoxiang.fan, maria.pilliguacosta, corentin.dumery, oliver.heinimann, pascal.fua\}@epfl.ch}
}

\newcommand{\method}{SGAR\xspace}
\newcommand{\methodDV}{SGAR~(DejaView)\xspace}
\newcommand{\methodVO}{SGAR~(VGGT-$\Omega$)\xspace}

\usepackage{pifont}

\definecolor{bestcol}{HTML}{a4ff94}
\definecolor{midcol}{HTML}{d2f894}
\definecolor{secondcol}{HTML}{fff194}
\newcommand{\best}[1]{\cellcolor{bestcol!80}#1}
\newcommand{\midrank}[1]{\cellcolor{midcol!80}#1}
\newcommand{\second}[1]{\cellcolor{secondcol!80}#1}
\newcommand{\nores}[1]{\textcolor{gray}{\textsc{#1}}}

\begin{document}
\input{sec/0_teaser}
{\renewcommand{\thefootnote}{}\footnotetext{$^{*}$Equal contribution.}}
\input{sec/0_abstract}
\input{sec/1_intro}
\input{sec/2_related_work}
\input{sec/3_methodology}
\input{sec/4_results}
\input{sec/5_conclusion}
{
    \small

\input{references.bbl}
}

\end{document}

%% file: preamble.tex
\input{defs}

%% file: defs.tex
\newif\ifdraft
\draftfalse

\definecolor{afcolor}{HTML}{00838F}
\definecolor{cdcolor}{HTML}{2D78C8}
\definecolor{mpccolor}{HTML}{8052B0}
\definecolor{ohcolor}{HTML}{B74990}
\definecolor{pfcolor}{HTML}{CA3B3B}
\definecolor{darkgreen}{RGB}{0,100,0}
\definecolor{darkred}{RGB}{140,0,0}

\ifdraft
  \newcommand{\authornote}[3]{{~\color{#2}\textbf{#1: #3}}}

  \newcommand{\deletion}[1]{{\color{red}[#1]}}

  \newcommand{\TODO}[1]{\textbf{\color{red}[TODO: #1]}}
\else
  \newcommand{\authornote}[3]{}

  \newcommand{\deletion}[1]{}

  \newcommand{\TODO}[1]{}
\fi

%% file: sec/0_teaser.tex
\twocolumn[{%
\renewcommand\twocolumn[1][]{#1}%
\maketitle
\begin{center}
    \centering
    \captionsetup{type=figure}
    \includegraphics[width=\textwidth]{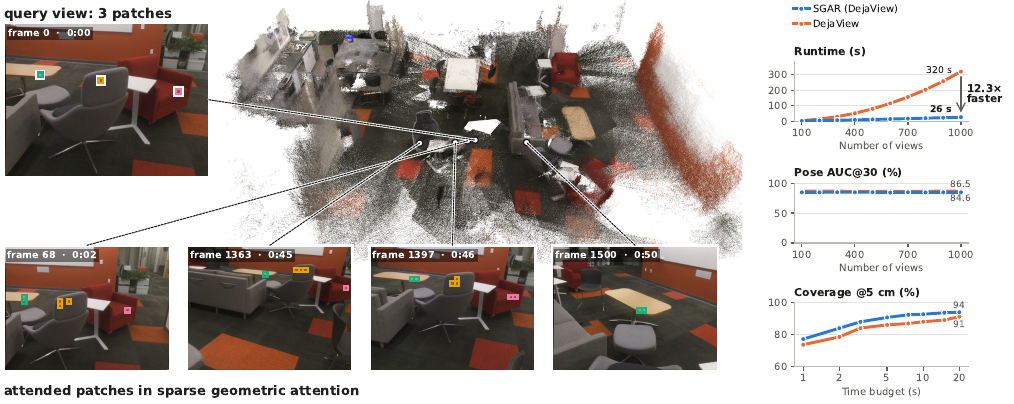}
    \captionof{figure}{\method{} replaces the dense global attention of feed-forward reconstruction with a sparse one: besides a few special tokens, each token attends only to the tokens that the model predicts to lie near it in 3D, which makes inference fast. Three tokens of the query view are marked in color, and the other views show the patches the model predicts for them. On this scene with $1{,}000$ views, \method{} takes $26.1$\,s against $319.6$\,s for DejaView on one Nvidia A100 GPU, at an AUC@30 of $84.6$ against $86.5$.
    }
    \label{fig:teaser}
\end{center}
\vspace{1em}
}]

%% file: sec/0_abstract.tex
\begin{abstract}
Feed-forward models have been very successful in 3D reconstruction with ad hoc attention designs that relate all view patches densely. However, the cost of this dense global attention grows quadratically with the number of views, preventing the model from applying to larger-scale reconstruction settings. In this paper, we argue that most dense attention computations are unnecessary from a classical structure-from-motion perspective: 3D structures stem from geometrically matching regions in 2D. We bring this principle to learning-based feed-forward reconstruction with our Sparse Geometric Attention Reconstruction (\method{}) method. In \method{}, each patch token attends to only a small set of patch tokens from other views, chosen by a lightweight learned key selection module trained to predict cross-view 3D proximity of tokens. Thus, dense global attention is replaced by much more efficient geometric-similarity-guided sparse attention. \method{} is a general method that can be applied to different feed-forward models by fine-tuning. In our experiments, we show that \method{} is the fastest among both state-of-the-art dense and sparse methods and the most accurate sparse one. It runs up to 12× faster than dense attention at 1,000 views with stable pose accuracy where existing sparse methods fail, and shows the best performance in reconstructing more complete geometries on the same time budget.

\end{abstract}

%% file: sec/1_intro.tex
\section{Introduction}
\label{sec:intro}

\input{fig/oracle}

Multi-view scene reconstruction is a long-standing problem in computer vision. Classical Structure-from-Motion (SfM) solutions treat this as a geometric estimation problem and solve it with a multi-stage pipeline built on projective geometry and robust optimization. Although effective, SfM methods are based on complex pipelines and produce only sparse reconstructions~\cite{OpenMVS26,Moulon16,Tola12,Pix4D,Schoenberger16a,Pan24a}. Recent feed-forward models such as VGGT~\cite{Wang25} and DejaView~\cite{Burzio26} replace it by a single network that predicts cameras and geometry for all views in a single forward pass, matching or even surpassing the earlier methods in a fraction of the time when the views overlap substantially. At the core of these methods is global attention: The tokens of all views are processed jointly, so that every view is interpreted in the context of every other. This is what makes these approaches so powerful, but also what makes them expensive as the cost of global attention grows quadratically with the total number of views. For example, in DejaView~\cite{Burzio26}, the time cost grows from a few seconds in $100$ views to more than $300$\,s in $1{,}000$ views on one Nvidia A100 GPU. This severely restricts the applicability of feed-forward models to time-sensitive applications, such as robot navigation, and to providing a fast initialization for optimization-based reconstruction pipelines.

However, much of this computation is redundant. As shown in Fig.~\ref{fig:oracle}, global attention is highly sparse even in densely trained models: When each query of DejaView attends only to a small fraction of the oracle keys determined based on ground truth geometry, the model retains most of the pose accuracy without any retraining. Many recent works leverage this observation to accelerate VGGT-style models, mostly without retraining. The strategies include merging similar tokens~\cite{Shen26,Chen26}, compressing each view into a few descriptors~\cite{Wang26c}, test-time training~\cite{Elflein26,Jin26}, or estimating the attention map at a coarse resolution and keeping only its strongest blocks~\cite{Wang26d,Deng26,Sun26,Ren26}. These methods infer the relevant keys from an approximation of the attention itself, rather than from the geometric relations between views on which reconstruction depends. As a consequence, their accuracy degrades significantly compared to dense methods, or they run out of memory on long sequences.

In this work, we revisit this problem from the perspective of classical Structure-from-Motion, in which views are related through sparse 2D correspondences. We hypothesize that, apart from a few tokens that anchor the global coordinate frame, the keys a patch token requires are the tokens that observe the same 3D surface region in other views. Key selection then becomes a geometric-feature retrieval problem, which can be supervised directly with ground-truth geometry and solved without computing dense attention.

Building on this insight, we propose \method{}, which replaces the dense global attention of feed-forward reconstruction models with geometry-guided sparse attention (Fig.~\ref{fig:teaser}). A lightweight projection, trained with a contrastive loss on ground-truth geometry, maps patch tokens into an embedding space in which tokens observing the same surface lie close together. At inference, these embeddings are clustered, and each patch token attends only to the patches of its nearest clusters, together with a small set of tokens that provide global context. Since the number of keys per token grows much more slowly than the total number of tokens, the cost of global attention becomes sub-quadratic in the number of views. \method{} can be applied to different dense attention models such as DejaView~\cite{Burzio26} and VGGT-$\Omega$~\cite{Wang26b} by fine-tuning pretrained weights and requires no ground-truth geometry at inference.

Experiments on six datasets show that \method{} is the fastest among state-of-the-art dense and sparse methods, running up to $12\times$ faster than dense models at $1{,}000$ views and keeping accuracy comparable to dense models in all scenarios. Because of its high efficiency, we also show that \method{} can handle many more images and cover much more extended scenes for a given computational budget.

\noindent In summary, our contributions are as follows.
\begin{itemize}
    \item We show that global attention in feed-forward reconstruction is highly sparse, and that the keys a query token requires can be predicted by their geometric similarities.
    \item We propose \method{}, a geometry-guided sparse attention mechanism whose keys are selected by a contrastively learned embedding, obtained by fine-tuning a pretrained dense model.
    \item We demonstrate that \method{} achieves the best trade-off between efficiency and accuracy, scaling to $1{,}000$ views on a single GPU, and reconstructing the most complete geometries at the same given time budget.
\end{itemize}
Code and models will be released.

%% file: fig/oracle.tex
\begin{figure}[t]
    \centering
    \includegraphics[width=\linewidth]{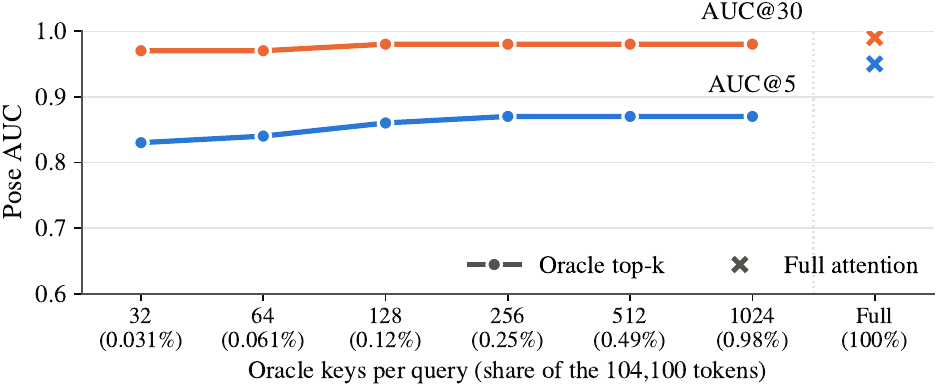}
    \caption{
    \textbf{Dense attention is redundant.} By selecting oracle top-k keys based on ground-truth geometry, most of the pose accuracy can be retained even when 99\% of attention weights are zeroed out. Evaluating DejaView~\cite{Burzio26} on an NRGBD~\cite{Azinovic22} scene.
    }
    \label{fig:oracle}
\end{figure}

%% file: sec/2_related_work.tex
\section{Related Work}
\label{sec:related}

\paragraph{Feed-forward 3D reconstruction.}
Structure-from-Motion pipelines~\cite{OpenMVS26,Moulon16,Tola12,Pix4D,Schoenberger16a} estimate poses and structure in stages, from keypoint matching to bundle adjustment, and learned matchers~\cite{Sarlin20,Sun21a,Lindenberger23} have made each stage more robust. DUSt3R~\cite{Wang24} and MASt3R~\cite{Leroy24} showed that a network can instead predict aligned pointmaps directly, but they operate on two images at a time and need an optimization to merge the pairwise outputs. VGGT~\cite{Wang25} and the models that followed it~\cite{Wang25b,Wang26b,Yang25,Lin26} read an arbitrary number of images at once, interleaving attention within each image and across all of them, and output cameras, depth and points without any optimization. DejaView~\cite{Burzio26} reaches comparable accuracy with far fewer parameters by applying one shared block recurrently. Streaming models~\cite{Wang25d,Wang25c} take a different route and ingest the images one by one, updating a persistent memory.
For the models that process all views jointly, the cost of dense global attention grows quadratically with the number of views. However, we observe that dense attention is highly redundant (Fig.~\ref{fig:oracle}), and \method{} therefore restricts attention to sparse, geometrically similar tokens.

\paragraph{Sparse attention.}
A growing body of work accelerates VGGT-style models, most of it without retraining. Token merging methods, following token merging for vision transformers~\cite{Bolya23}, reduce the tokens that enter global attention, merging near-duplicate patches~\cite{Shen26} or low-confidence ones ranked by a distilled predictor~\cite{Chen26}. FlashVGGT~\cite{Wang26c} compresses each view into a few descriptor tokens and turns global attention into cross-attention to them. VGG-T$^3$~\cite{Elflein26} and ZipMap~\cite{Jin26} distill the key-value set into a fixed-size network by test-time training. AVGGT~\cite{Sun26} turns the early global layers into frame attention and subsamples the keys of the remaining ones on the patch grid, and Lite3R~\cite{Zhang26} distills a sparse linear attention into the frozen backbone. Block-Sparse~\cite{Wang26d} pools the keys into blocks and reads only those that hold the attention mass, SAF3R~\cite{Deng26} fits a sparse pattern to each head, and Speed3R~\cite{Ren26} pairs a compressed branch with a sparse one that attends to the blocks it selects. Most of these methods approximate dense attention at the granularity of blocks or groups of tokens at the risk of over-simplification.  \method{} instead selects the exact keys that are meaningful for every token individually from a geometric point of view.

Restricting attention to a subset of keys is a long-standing remedy in the broader literature too. Early methods fix the pattern in advance, with local windows and strided or global tokens~\cite{Child19,Beltagy20,Zaheer20,Liu21a}, while content-based methods group queries and keys by hashing~\cite{Kitaev20} or clustering~\cite{Roy21}, or retrieve the nearest keys from a large memory~\cite{Wu22}. We investigate the problem in a different context of geometric reconstruction and learn key selection by training geometric representations.

\paragraph{Learned geometric embeddings.}
Finding correspondences between images rests on learning geometric embeddings of points whose similarity reflects whether two locations observe the same 3D point. Learned geometric embeddings are trained with metric or contrastive objectives on ground-truth correspondences~\cite{Choy16a}. Modern matchers reason over both images with attention, on sparse keypoints~\cite{Sarlin20,Lindenberger23,tyszkiewicz2020disk} or densely~\cite{Sun21a,edstedt2024roma}. MASt3R~\cite{Leroy24} brings matching into feed-forward reconstruction with a head trained by an InfoNCE loss~\cite{Oord18}, whose descriptors are matched after the forward pass. Supervised contrastive learning~\cite{Khosla20} extends InfoNCE to several positives per anchor, as when a patch overlaps several patches of another view. \method{} uses the same idea as a supervision signal during training: a multi-positive InfoNCE loss pulls together the embeddings of corresponding patches and pushes away all other patches. However, our learning objective is not to find point correspondences, but to learn patch-level geometric embeddings that are never matched explicitly. The geometric similarities will be used to sparsify global attention with reconstruction as the final objective.

%% file: sec/3_methodology.tex
\input{fig/method}

\section{Method}
\label{sec:method}

We introduce \method{}, a sparse-attention reconstruction scheme in which attention across views is restricted to patches that correspond to the same 3D neighborhood.
This makes its cost sub-quadratic in the number of views as opposed to quadratic in dense attention. It rests on two things: a learned key selection that clusters the tokens from different views likely to be in 3D proximity, and a sparse global attention mechanism in which every patch token attends in full only to the patches of its own clusters and the 3 nearest clusters, as well as to the rest of the scene through a few global tokens and cluster summaries. \method{} can be readily applied to any dense attention models such as DejaView~\cite{Burzio26} and VGGT-$\Omega$~\cite{Wang26b}, by replacing its global attention with our sparse ones and fine-tuning. This yields the complete architecture of Fig.~\ref{fig:method}, where DejaView is used as an example base architecture.  We define the key selection process in Sec.~\ref{sec:index-learning}, describe our sparse attention mechanism in  Sec.~\ref{sec:sparse}, and present our training procedure in Sec.~\ref{sec:training}.

\paragraph{Notation.}
The input is a set of $V$ RGB images of the same scene $(I_i)_{i=1}^{V}$. Each image is patchified into $P$ tokens and augmented with one camera token and four register tokens~\cite{Darcet24}, which we jointly call the \emph{special} tokens, giving $T = P + 5$ tokens of dimension $C$ per view. The tokens are refined by a number of alternations of frame attention and global attention. We write $N = VP$ for the number of patch tokens, index a patch token by $i \in \{1, \dots, N\}$, and write $\mathcal{S}$ for the $5V$ special tokens and $\mathcal{R}$ for the $P$ patch tokens of the first view, which is the reference view in whose camera frame the model predicts the scene.

\subsection{Learning key selection}
\label{sec:index-learning}

Key selection aims to identify the relevant patches that will restrict attention in \cref{sec:sparse}. We first compute geometric embeddings for every patch token with a learned projection $W_{\text{adj}}$, where $W_{\text{adj}} x_i$ and $W_{\text{adj}} x_j$ are close in feature space if and only if $x_i$ and $x_j$ belong to the same neighborhood in 3D. We then consider these embeddings across all views, and cluster them in feature space. By design, the resulting clusters correspond to close 3D regions and will be used to restrict attention computation in~\cref{sec:sparse}.

\paragraph{Geometric embedding.}
The shared block carries a linear \emph{adjacency projection} $W_{\text{adj}} \in \mathbb{R}^{C \times C}$, applied to the patch tokens after frame attention, as is common in contrastive learning~\cite{Chen20a,Radford21}. Its outputs are centered over the patches of the sample and normalized,
\begin{equation}
  e_i = \frac{W_{\text{adj}} x_i - \mu}{\lVert W_{\text{adj}} x_i - \mu \rVert}, \qquad
  \mu = \frac{1}{N} \sum_{j=1}^{N} W_{\text{adj}} x_j ,
  \label{eq:embedding}
\end{equation}
so that two patches are compared by the cosine $\langle e_i, e_j \rangle$ of their embeddings, which is a normalized Mahalanobis inner product with a metric learned as in~\cite{Weinberger08}. Centering removes the component shared by all patches of a scene, which would otherwise dominate the cosines and pull every embedding into one direction. The embedding is trained so that patches observing the same surface point have nearby embeddings, with the contrastive loss defined below.

\paragraph{Contrastive loss.}
We train $W_{\text{adj}}$ with a multi-positive InfoNCE~\cite{Oord18,Khosla20} on the same cosine that the $k$-means clusters by. We sample a set $\mathcal{Q}$ of $1024$ query patches with valid ground truth and keep those with at least one corresponding patch. For a query $i$, let $\mathcal{P}_i$ be its corresponding patches and $\mathcal{A}_i \supseteq \mathcal{P}_i$ the candidates: all patches of the other views, except those within five patches of a corresponding patch that do not correspond themselves, whose status is an artefact of the patch grid. With $e^{(l)}_i$ the embedding of Eq.~\eqref{eq:embedding} at layer $l$, the loss is
\begin{equation}
  \mathcal{L}_{\text{con}} = -\frac{1}{L\,|\mathcal{Q}|} \sum_{l=1}^{L} \sum_{i \in \mathcal{Q}}
  \log \frac{\sum_{j \in \mathcal{P}_i} \exp\!\big(\tau \langle e^{(l)}_i, e^{(l)}_j \rangle\big)}
            {\sum_{j \in \mathcal{A}_i} \exp\!\big(\tau \langle e^{(l)}_i, e^{(l)}_j \rangle\big)} ,
  \label{eq:infonce}
\end{equation}
averaged over the samples of the batch, where $\tau$ is a learned temperature initialised at $20$ and capped at $100$. The projection reads the tokens through a stop-gradient, so $\mathcal{L}_{\text{con}}$ trains $W_{\text{adj}}$ and $\tau$ alone, while the reconstruction losses train the backbone. The clustering itself is not differentiable, the backbone learns to use the clusters it is given.

To generate the ground truth, we call two patches of different views \emph{corresponding} when they observe the same surface. Patch $i$ of view $u$ corresponds to patch $j$ of view $v$ if a pixel of $i$, unprojected with its ground-truth depth and reprojected into $v$, lands within one pixel of a pixel of $j$ whose ground-truth depth matches to within $5\%$.

\paragraph{Key selection by clustering.}
Instead of ranking the keys of every query individually, we group all $N$ patches at once. The embeddings $\{e_i\}$ are clustered by spherical $k$-means~\cite{Dhillon01}, seeded by $k$-means++~\cite{Arthur07}, into $M = \sqrt{N}$ clusters: every patch is assigned to the centroid with the highest cosine similarity, and every centroid is set to the normalized mean of its patches. Each cluster $c$ then selects the $m = 4$  clusters $\mathcal{N}(c)$ whose centroids are most similar to its own, itself first. A patch $i$ of cluster $c(i)$ attends to all patches of these clusters,
\begin{equation}
  \mathcal{I}(i) = \{\, j : c(j) \in \mathcal{N}(c(i)) \,\} .
  \label{eq:index}
\end{equation}
The nearest clusters cover the patches at the boundary of a cluster, and the surfaces that $k$-means splits into several clusters, which a single cluster would miss. The selection is made per cluster rather than per patch, so that all queries of a cluster share their keys; this is what lets the attention run as dense tiles on tensor cores (Sec.~\ref{sec:sparse}). The clustering is redone at every layer, so that it follows the model's evolving estimate of the scene.

\subsection{Sparse global attention}
\label{sec:sparse}
With the learned key selection, dense attention can be sparsified. We replace a dense model's global attention with a few residual updates that reuse the model's global attention weights. First, we let the special tokens attend densely to all $VT$ tokens to preserve a global channel that the clusters do not restrict. Each patch token is then updated by four attentions computed from one projection $(q, k, v)$ of the patch tokens, each with its own softmax, whose outputs are summed and passed through the output projection $W_o$:
\begin{equation}
  x_i \leftarrow x_i + W_o \!\!\sum_{\mathcal{K} \in \{\mathcal{S},\, \mathcal{R},\, \mathcal{M}(i),\, \mathcal{I}(i)\}} \!\!\!\mathrm{softmax}_{j \in \mathcal{K}}\Big(\frac{q_i^\top k_j}{\sqrt{D}} + b_j\Big)\, v_j .
  \label{eq:attn}
\end{equation}

where $D$ is the head dimension, and the four key sets in $\mathcal{K}$ are the special tokens $\mathcal{S}$, the patches $\mathcal{R}$ of the reference view, the patches $\mathcal{I}(i)$ of the nearest clusters, and one summary token $\mathcal{M}(i)$ for every cluster that is not in $\mathcal{I}(i)$.

\paragraph{Implementation and complexity.}
The patches are sorted by cluster, and every cluster is arranged to contain contiguous blocks that are padded to $128$, the tile of the attention kernel. Every tile of $128$ queries then belongs to a single cluster and attends to the whole tiles of closest clusters, so the attention over $\mathcal{I}(i)$ is a block-sparse FlexAttention~\cite{Dong25} whose mask we build directly from the block boundaries. A patch attends to $5V + P + 5\sqrt{N}$ keys on average, $5V$ special tokens, $P$ reference patches, $\sqrt{N}$ summaries and four clusters of $\sqrt{N}$ patches on average, instead of $VT$. Global attention thus costs $O\big(N(V + P + \sqrt{N})D\big)$ rather than $O\big((VT)^2 D\big)$, and each round of $k$-means costs $O(N^{1.5} C)$. At $1{,}000$ views of $972$ patches, a patch attends to about $11{,}000$ keys, about $1\%$ of the tokens.

\subsection{Training}
\label{sec:training}

We train \method{} by fine-tuning from a pretrained dense attention model (\eg DejaView, VGGT-$\Omega$). The adjacency projection and the temperature $\tau$ of $\mathcal{L}_\mathrm{con}$ are the only new parameters. The model is trained with the following loss:
\begin{equation}
  \mathcal{L} = \mathcal{L}_{\text{recon}} + \lambda_{\text{con}} \mathcal{L}_{\text{con}} ,
  \label{eq:loss}
\end{equation}
where $\mathcal{L}_{\text{recon}}$ are the dense model's point-map, depth, ray and camera losses on the ground truth, expressed in the camera frame of the reference view after scale normalisation. The second term $\mathcal{L}_{\text{con}}$ is the contrastive loss~\eqref{eq:infonce}, with $\lambda_{\text{con}} = 1$.

\paragraph{Training Details.}
We train on a mixture of $18$ public datasets that covers indoor and outdoor scenes, driving, aerial and object-centric captures, both real and synthetic: ScanNet~\cite{Dai17b}, Hypersim~\cite{Roberts21}, Replica~\cite{Straub19}, Taskonomy~\cite{Zamir18}, CA-1M~\cite{Lazarow25}, Aria Synthetic Environments~\cite{Avetisyan24}, MegaDepth~\cite{Li18o}, BlendedMVS~\cite{Yao20b}, CO3D~\cite{Reizenstein21}, renderings of TRELLIS 3D assets~\cite{Xiang25a}, Mapillary Metropolis~\cite{Mapillary21b}, Virtual KITTI~2~\cite{Cabon20}, TartanAir and its V2 release~\cite{Wang20n}, MVS-Synth~\cite{Huang18b}, UnrealStereo4K~\cite{Tosi21}, Spring~\cite{Mehl23} and Kubric-4D~\cite{Greff22,VanHoorick24}.
A sample of $V$ views is drawn only from scenes with at least $V$ frames, without repeated frames. In training runs, the view samples hold $2$ to $32$ views. The aspect ratio of images is drawn per batch between $1{:}3$ and $1{:}1$, and receives color, blur, and zoom augmentation. Since the dense model's representation is to be fine-tuned rather than relearned, the pretrained weights are trained at a peak learning rate of $10^{-5}$ and the new parameters at $10^{-4}$, with AdamW. We train on 4 H100 GPUs in bfloat16, with gradient checkpointing of every layer, and evaluate the checkpoint after $50$k steps.

%% file: fig/method.tex
\begin{figure*}
    \centering
    \includegraphics[width=1\linewidth]{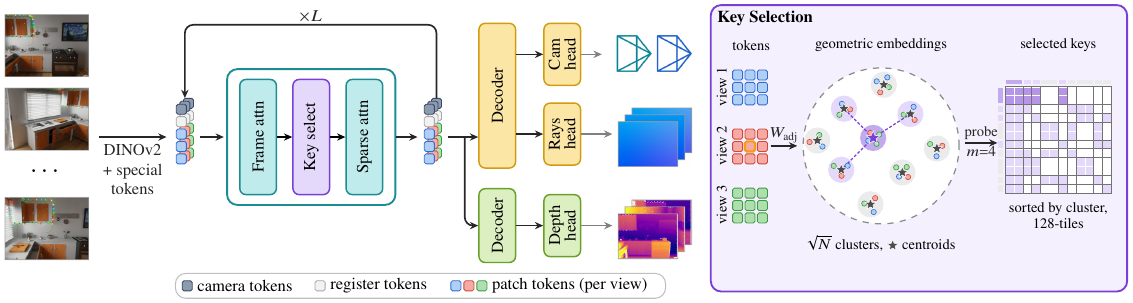}
    \caption{\textbf{Overview of \method{}.} We replace the global dense attention of DejaView~\cite{Burzio26} with our sparse geometric attention as an example. The images are encoded by DINOv2 and augmented with camera and register tokens. A single shared block, unrolled over its $L$ steps, runs frame attention, key selection and sparse attention. In sparse attention, each patch attends to the special tokens of all views, the patches of the reference view, a summary of each cluster, and all the patches of the nearest clusters, so that its cost grows near-linearly with the number of views. \textbf{Key Selection} projects the patch tokens by $W_{\text{adj}}$ into a geometric embedding and clusters the embeddings by spherical $k$-means into $\sqrt{N}$ clusters, re-clustered at every step. Each cluster probes its $m = 4$ nearest clusters, itself first. With the patches sorted by cluster, the selected keys form a block-sparse mask over tiles of $128$ tokens. A contrastive loss pulls together the embeddings of patches that observe the same surface in the ground-truth geometry and pushes the others apart.
    }
    \label{fig:method}
\end{figure*}

%% file: sec/4_results.tex
\section{Results}

\begin{table*}[t]
\centering
\caption{\textbf{Camera pose estimation.} \methodVO and \methodDV are the fastest models at every sequence length and the most accurate efficient ones, close to their dense counterparts. AUC in \%, time in s per scene, Params in millions, Mem in peak GiB on Mip-NeRF~360; \nores{oom}: out of memory; \nores{fail}: kernel compile error; --: not available. Green, yellow-green, yellow: first, second, third; ties share a place.
}
\label{tab:main}
\scriptsize
\setlength{\tabcolsep}{3.5pt}
\begin{tabular}{l r rrrr rr r rrr rrr r}
\toprule
 & & \multicolumn{7}{c}{Short ($<100$)} & \multicolumn{3}{c}{Mid-range (125--311)} & \multicolumn{3}{c}{Long (1{,}000)} & \\
\cmidrule(lr){3-9}\cmidrule(lr){10-12}\cmidrule(lr){13-15}
 & & \multicolumn{4}{c}{AUC@30} & \multicolumn{2}{c}{Mean AUC} & & \multicolumn{2}{c}{Mip360 AUC} & & \multicolumn{2}{c}{ScanNet AUC} & & \\
\cmidrule(lr){3-6}\cmidrule(lr){7-8}\cmidrule(lr){10-11}\cmidrule(lr){13-14}
Method & Params & 7-Scenes & ETH3D & LaMAR & NRGBD & @5 & @30 & Time & @5 & @30 & Time & @5 & @30 & Time & Mem \\
\midrule
VGGT~\cite{Wang25} & 1158 & 81.0 & 70.6 & 27.3 & 97.2 & 40.2 & 69.0 & 6.98 & 74.5 & 94.9 & 54.1 & \nores{oom} & -- & -- & 75.3 \\
$\pi^3$X~\cite{Wang25b} & 1360 & 82.0 & 83.7 & \second{59.1} & \midrank{98.6} & \second{51.8} & 80.8 & 3.82 & \nores{oom} & -- & -- & \nores{oom} & -- & -- & -- \\
DA3~\cite{Lin26} & 1356 & 82.8 & 83.6 & 47.7 & \best{99.8} & \midrank{53.5} & 78.5 & 3.81 & \best{90.9} & \best{98.3} & 24.3 & \nores{oom} & -- & -- & 39.4 \\
VGGT-$\Omega$~\cite{Wang26b} & \second{1144} & \best{88.2} & \best{90.1} & \best{59.7} & \midrank{98.6} & \best{57.9} & \best{84.2} & 2.27 & \midrank{86.3} & \second{97.4} & 14.2 & \midrank{27.9} & \midrank{85.3} & 303.8 & \best{22.9} \\
DejaView~\cite{Burzio26} & \best{117} & \second{83.7} & \second{87.8} & \midrank{59.2} & \midrank{98.6} & 48.6 & \midrank{82.3} & 2.10 & \second{85.6} & \midrank{97.5} & 14.1 & \best{31.1} & \best{86.5} & 319.6 & 27.4 \\
\midrule
Block-Sparse~\cite{Wang26d} & 1158 & 80.8 & 70.5 & 27.2 & 97.1 & 39.8 & 68.9 & 4.31 & 73.3 & 94.7 & 28.9 & \nores{oom} & -- & -- & 26.6 \\
FastVGGT~\cite{Shen26} & 1158 & 79.5 & 66.4 & 26.6 & 96.2 & 35.3 & 67.2 & 3.47 & 72.3 & 94.5 & 16.5 & 0.0 & 3.2 & 257.0 & 26.3 \\
SAF3R~\cite{Deng26} & 1256 & 79.7 & 56.3 & 21.1 & 95.6 & 32.0 & 63.2 & 3.55 & 74.3 & 94.9 & 18.0 & \nores{fail} & -- & -- & 26.9 \\
AVGGT~\cite{Sun26}, $\sigma{=}2$ & 1158 & 80.2 & 63.3 & 21.4 & 96.7 & 37.3 & 65.4 & 3.52 & 72.2 & 93.8 & 20.6 & \nores{oom} & -- & -- & 76.4 \\
AVGGT~\cite{Sun26}, $\sigma{=}4$ & 1158 & 79.6 & 59.9 & 20.7 & 96.4 & 35.6 & 64.1 & 2.76 & 70.4 & 93.2 & 13.6 & \nores{oom} & -- & -- & 76.0 \\
FlashVGGT~\cite{Wang26c} & 1158 & 77.6 & 57.5 & 16.7 & 93.1 & 29.1 & 61.2 & 2.12 & 58.8 & 90.2 & 8.1 & \nores{oom} & -- & -- & 28.0 \\
Co-Me~\cite{Chen26} & 1191 & 75.4 & 56.9 & 9.8 & 87.5 & 21.8 & 57.4 & 2.09 & 36.6 & 81.4 & 11.3 & 3.9 & 35.4 & \second{203.4} & \midrank{23.0} \\
StreamVGGT~\cite{Zhuo25} & 1256 & 75.7 & 54.7 & 17.6 & 88.9 & 25.9 & 59.2 & 13.95 & 39.1 & 79.3 & 109.6 & \nores{oom} & -- & -- & 42.7 \\
VGG-T$^3$~\cite{Elflein26} & 1192 & 68.1 & 44.0 & 14.1 & 88.2 & 20.1 & 53.6 & 3.80 & 15.2 & 65.1 & 11.1 & \nores{oom} & -- & -- & 49.0 \\
ZipMap~\cite{Jin26} & 1334 & 80.5 & 73.4 & 39.6 & 96.5 & 39.8 & 72.5 & \second{1.81} & 61.6 & 90.8 & \second{5.5} & \nores{oom} & -- & -- & 32.3 \\
\midrule
\textbf{\methodVO{}} & \textbf{1164} & \textbf{\midrank{84.0}} & \textbf{\midrank{88.6}} & \textbf{54.5} & \textbf{\second{97.7}} & \textbf{51.3} & \textbf{\second{81.2}} & \textbf{\midrank{1.65}} & \textbf{81.9} & \textbf{96.5} & \textbf{\midrank{5.4}} & \textbf{23.5} & \textbf{83.5} & \textbf{\midrank{40.5}} & \textbf{\second{23.7}} \\
\textbf{\methodDV{}} & \textbf{\midrank{118}} & \textbf{82.3} & \textbf{83.6} & \textbf{49.7} & \textbf{97.3} & \textbf{42.7} & \textbf{78.2} & \textbf{\best{1.32}} & \textbf{78.3} & \textbf{96.0} & \textbf{\best{4.1}} & \textbf{\second{27.1}} & \textbf{\second{84.6}} & \textbf{\best{26.1}} & \textbf{28.5} \\
\bottomrule
\end{tabular}
\end{table*}

\begin{figure*}[t]
    \centering
    \includegraphics[width=\textwidth]{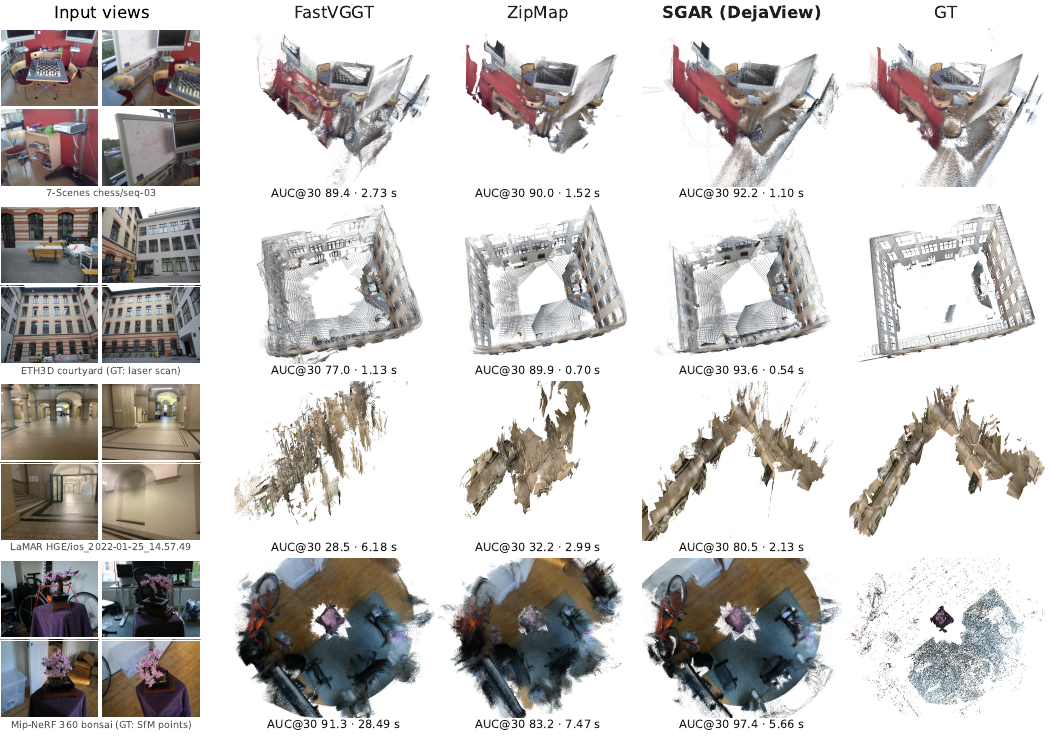}
    \caption{\textbf{Qualitative comparison.} Point clouds predicted by FastVGGT, ZipMap and \methodDV, with the ground truth, on one scene of each of 7-Scenes (chess/seq-03), ETH3D (courtyard, laser scan), LaMAR (HGE) and Mip-NeRF~360 (bonsai, SfM points). Left: four of the input views. Below each reconstruction: the AUC@30 of its cameras. \methodDV keeps the structure of the scene where the other efficient models fragment it, most visibly on the long LaMAR trajectory.
    }
    \label{fig:qualitative}
\end{figure*}

\begin{table}[t]
\centering
\caption{\textbf{Reconstruction at a fixed time budget.} Coverage @5\,cm (\%): share of the ground-truth surface of ScanNet scene0770\_00 within $5$\,cm of a predicted point after ICP alignment, with each model given as many views as fit in the budget.
Of each column, over all models, green marks the best value, yellow-green the second and yellow the third. Our model can recover the most complete geometries at all time budgets.
}
\label{tab:budget}
\scriptsize
\setlength{\tabcolsep}{2.4pt}
\begin{tabular}{l rrrrrrrr}
\toprule
 & \multicolumn{8}{c}{Time budget (s)} \\
\cmidrule(lr){2-9}
Method & 1 & 2 & 3 & 5 & 7.5 & 10 & 15 & 20 \\
\midrule
VGGT~\cite{Wang25} & 59.5 & 69.9 & 73.2 & 78.6 & 81.2 & 84.4 & 87.0 & 88.3 \\
$\pi^3$X~\cite{Wang25b} & 55.9 & 67.5 & 73.5 & 80.1 & 80.6 & 81.9 & 84.9 & 85.5 \\
DA3~\cite{Lin26} & 59.8 & 74.4 & 75.9 & 83.6 & 84.7 & 87.3 & 88.8 & 89.9 \\
VGGT-$\Omega$~\cite{Wang26b} & 71.0 & 79.6 & 82.5 & 86.4 & 88.7 & 90.7 & 90.9 & 92.4 \\
DejaView~\cite{Burzio26} & \second{73.8} & 78.6 & 84.2 & 86.2 & 87.0 & 88.2 & 89.2 & 91.3 \\
\midrule
Block-Sparse~\cite{Wang26d} & 60.4 & 73.1 & 77.9 & 82.7 & 85.2 & 87.0 & 89.1 & 90.4 \\
FastVGGT~\cite{Shen26} & 41.5 & 71.2 & 79.9 & 84.9 & 87.4 & 89.2 & 90.0 & 91.8 \\
SAF3R~\cite{Deng26} & 11.6 & 71.2 & 80.9 & 87.7 & 89.0 & 90.9 & 92.2 & 93.2 \\
AVGGT~\cite{Sun26}, $\sigma{=}2$ & 59.9 & 75.2 & 79.5 & 84.5 & 86.4 & 89.6 & 91.1 & 92.1 \\
AVGGT~\cite{Sun26}, $\sigma{=}4$ & 64.5 & 79.2 & 81.5 & 87.0 & 87.5 & 90.2 & 91.6 & \second{93.6} \\
FlashVGGT~\cite{Wang26c} & 55.7 & 78.0 & 84.3 & \second{88.9} & \second{90.7} & \second{91.2} & \second{92.9} & 93.4 \\
Co-Me~\cite{Chen26} & 59.0 & 73.1 & 80.9 & 83.4 & 86.8 & 87.3 & 90.2 & 91.4 \\
StreamVGGT~\cite{Zhuo25} & 11.6 & 11.2 & 39.1 & 57.8 & 63.5 & 69.5 & 74.2 & 73.1 \\
VGG-T$^3$~\cite{Elflein26} & 9.2 & 60.5 & 62.3 & 70.4 & 84.3 & 82.2 & 79.2 & 74.2 \\
ZipMap~\cite{Jin26} & 65.0 & \second{79.8} & \second{85.2} & 88.1 & 89.3 & 90.1 & 92.2 & 92.9 \\
\midrule
\textbf{\methodVO{}} & \textbf{\best{80.7}} & \textbf{\best{89.4}} & \textbf{\best{90.6}} & \textbf{\best{93.3}} & \textbf{\best{94.0}} & \textbf{\best{95.2}} & \textbf{\best{95.5}} & \textbf{\best{95.9}} \\
\textbf{\methodDV{}} & \textbf{\midrank{77.3}} & \textbf{\midrank{84.1}} & \textbf{\midrank{88.1}} & \textbf{\midrank{90.8}} & \textbf{\midrank{92.5}} & \textbf{\midrank{92.8}} & \textbf{\midrank{93.8}} & \textbf{\midrank{94.1}} \\
\bottomrule
\end{tabular}
\end{table}

\begin{table}[t]
\centering
\caption{\textbf{Depth estimation.} Depth on the short benchmark ($K = 100$ views per scene) for the three datasets with ground-truth depth. The predicted depth is aligned to the ground truth with one median scale per sequence, so it must be consistent across views. $\delta_{1.25}$: share of pixels with $\delta{<}1.25$, in per cent (higher is better). Mean is over the three datasets, each weighted equally. Of each column, over all models, green marks the best value, yellow-green the second and yellow the third; tied values share a place.}
\label{tab:depth}
\scriptsize
\setlength{\tabcolsep}{4pt}
\begin{tabular}{l rrrr}
\toprule
 & \multicolumn{4}{c}{$\delta_{1.25}$~$\uparrow$} \\
\cmidrule(lr){2-5}
Method & 7-Scenes & LaMAR & NRGBD & Mean \\
\midrule
VGGT~\cite{Wang25} & 95.4 & 81.1 & 99.5 & 92.0 \\
$\pi^3$X~\cite{Wang25b} & \best{98.5} & \best{94.0} & 99.6 & \best{97.3} \\
DA3~\cite{Lin26} & 97.5 & 91.6 & 99.5 & 96.2 \\
VGGT-$\Omega$~\cite{Wang26b} & 97.4 & \second{91.8} & \best{100.0} & \second{96.4} \\
DejaView~\cite{Burzio26} & 97.6 & 87.3 & 99.3 & 94.8 \\
\midrule
Block-Sparse~\cite{Wang26d} & 95.4 & 81.0 & 99.5 & 92.0 \\
FastVGGT~\cite{Shen26} & 95.8 & 81.0 & 99.4 & 92.1 \\
SAF3R~\cite{Deng26} & 95.8 & 74.9 & 99.1 & 89.9 \\
AVGGT~\cite{Sun26}, $\sigma{=}2$ & 95.5 & 74.9 & 99.4 & 90.0 \\
AVGGT~\cite{Sun26}, $\sigma{=}4$ & 95.7 & 73.9 & 99.5 & 89.7 \\
FlashVGGT~\cite{Wang26c} & 96.1 & 72.9 & 99.5 & 89.5 \\
Co-Me~\cite{Chen26} & 94.0 & 73.9 & 95.7 & 87.9 \\
StreamVGGT~\cite{Zhuo25} & 94.7 & 60.3 & 99.1 & 84.7 \\
VGG-T$^3$~\cite{Elflein26} & 96.6 & 89.6 & 99.5 & 95.2 \\
ZipMap~\cite{Jin26} & 95.3 & \midrank{93.0} & \second{99.7} & 96.0 \\
\midrule
\textbf{\methodVO{}} & \textbf{\midrank{98.4}} & \textbf{91.1} & \textbf{\midrank{99.9}} & \textbf{\midrank{96.5}} \\
\textbf{\methodDV{}} & \textbf{\second{98.1}} & \textbf{76.5} & \textbf{99.3} & \textbf{91.3} \\
\bottomrule
\end{tabular}
\end{table}

\subsection{Experimental setup}
\label{sec:setup}

\paragraph{Benchmark.}
We evaluate zero-shot camera pose and depth estimation in three ranges of sequence length, on five public datasets that are not in our training data, $200$ scenes in total, and on one long capture from the ScanNet test split, which is held out from training. \emph{Short}: 7-Scenes~\cite{Shotton13c}, ETH3D~\cite{Schops17}, LaMAR~\cite{Sarlin22} and NRGBD~\cite{Azinovic22}, each scene subsampled at a fixed frame rate and evaluated on its first $K = 100$ views. \emph{Mid-range}: the $9$ scenes of Mip-NeRF~360~\cite{Barron22}, each on all of its $125$ to $311$ views ($194$ median); they are photo collections around an object rather than videos, so the views are far apart but all observe the same object. \emph{Long}: ScanNet~\cite{Dai17b} test scene0770\_00, a $3{,}383$-frame capture whose camera covers the largest floor area of the test set, swept from $K = 100$ to $1{,}000$ views spaced evenly over the capture.

\paragraph{Metrics.}
Following~\cite{Wang25}, the error of a view pair is the larger of its relative rotation and translation angular errors, and we report the area under the pair-error curve up to $30^\circ$ (AUC@30) and $5^\circ$ (AUC@5), averaged over the scenes of a dataset and then over datasets. Depth is evaluated on the short datasets with ground-truth depth, 7-Scenes, LaMAR and NRGBD: the predicted depth is aligned to the ground truth with one median scale per sequence, so that it must be consistent across views, and we report the share of pixels with $\delta{<}1.25$.

\paragraph{Baselines.}
We apply \method{} to two pretrained dense models of very different size and design, and evaluate both: \methodDV, fine-tuned from DejaView~\cite{Burzio26}, and \methodVO, fine-tuned from VGGT-$\Omega$~\cite{Wang26b}. We compare with the dense VGGT~\cite{Wang25}, $\pi^3$X~\cite{Wang25b}, Depth Anything~3 (DA3, giant)~\cite{Lin26}, VGGT-$\Omega$~\cite{Wang26b} and DejaView~\cite{Burzio26}. We also compare with ten efficient models: Block-Sparse~\cite{Wang26d}, FastVGGT~\cite{Shen26}, SAF3R~\cite{Deng26}, AVGGT~\cite{Sun26} at two sparsity levels $\sigma$, FlashVGGT~\cite{Wang26c}, Co-Me~\cite{Chen26}, the causal streaming StreamVGGT~\cite{Zhuo25}, and VGG-T$^3$~\cite{Elflein26} and ZipMap~\cite{Jin26} which replaces global attention with test-time training layers of linear cost. Each model runs with its released weights. All models run on one A100 80\,GB in bfloat16, and we time the full forward pass of a scene, excluding image loading and a warm-up run per input shape.

\begin{figure*}[t]
    \centering
    \includegraphics[width=\textwidth]{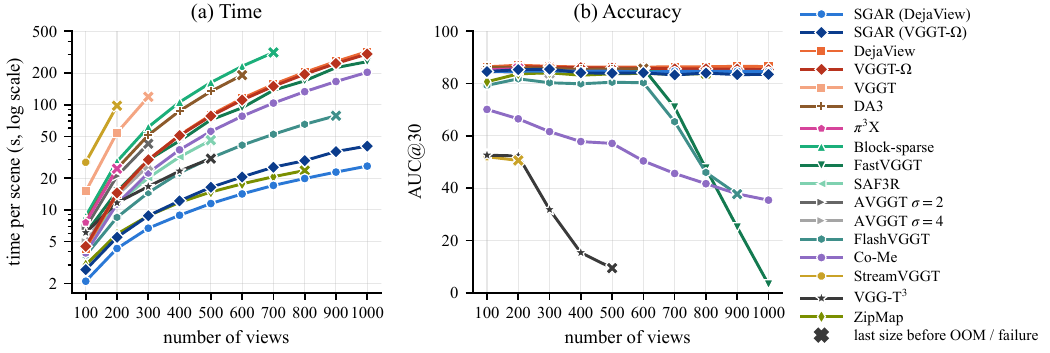}
    \caption{\textbf{Scaling with the number of views.} Time per scene (a, log scale) and AUC@30 (b) on the ScanNet test capture scene0770\_00, with $100$ to $1{,}000$ views evenly spaced over its $3{,}383$ frames, so that more views sample the same trajectory more densely. A line ends where the model runs out of memory on the A100 80\,GB or fails; the cross marks its last size.}
    \label{fig:views}
\end{figure*}

\subsection{Comparison with the state of the art}
\label{sec:comparison}

\paragraph{Accuracy and cost.}
Tab.~\ref{tab:main} compares all models in the three ranges of sequence length. Both variants are the fastest models in every range, and in every range they are also the most accurate of the efficient models. \method{} also greatly improves the efficiency of the pretrained dense attention models while maintaining comparable accuracies. On the long ScanNet sequence, \method{} runs up to $12\times$ faster than the dense models at close pose accuracy, whereas every other efficient model runs out of memory, fails, or loses most of its accuracy. We also compare the depth estimation results in Tab.~\ref{tab:depth}, our \methodVO is the most accurate efficient model, which matches the performance of VGGT-$\Omega$. We also provide qualitative results in Fig.~\ref{fig:qualitative} with comparison to FastVGGT and ZipMap, where our method shows both the best efficiency and point map quality.

\paragraph{Scaling with the number of views.}
Fig.~\ref{fig:views} provides testing results on the long ScanNet capture up to $1{,}000$ views.
 The time of both our variants grows far more slowly than that of the dense models, and their accuracy stays nearly flat. At $1{,}000$ views, \methodDV is $12\times$ faster than DejaView and VGGT-$\Omega$, while maintaining comparable accuracies. \methodVO is $7.5\times$ faster than VGGT-$\Omega$ at $1{,}000$ views and retaining most of its AUC@30. The other efficient models run out of memory or fail before $1{,}000$ views, or lose much of their accuracy.

\paragraph{Reconstruction at a fixed time budget.}
To measure what the speed buys, we also fix the inference time instead of the number of views for a time budget experiment as in Tab.~\ref{tab:budget}. We use the ScanNet capture of the long benchmark and budgets from $1$ to $20$\,s. For each model and budget, we find the largest number of views, spread evenly over the capture, whose forward pass fits in the budget. Every pixel's predicted depth is unprojected with the predicted cameras, aligned to the ground truth by the similarity that maps the predicted cameras onto the ground-truth ones, refined by a similarity ICP~\cite{Besl92,Umeyama91} onto the ground-truth surface, and voxelised at $1$\,cm. We report the reconstruction coverage @5\,cm, \ie the share of the ground-truth mesh within $5$\,cm of a predicted point.
Because of their efficiency, \methodDV and \methodVO both fit many more views than the corresponding dense models into the same time budget, and these extra views translate into a more complete reconstruction. Moreover, our method also compares favorably against all sparse models.

\begin{table}[t]
\centering
\caption{\textbf{Ablations.} An ablation study on the \methodDV model. AUC@30 (\%) and time (s per scene). $K = 100$: mean over the four short datasets; ScanNet: $1{,}000$ views.}
\label{tab:abl_index}
\footnotesize
\setlength{\tabcolsep}{3pt}
\begin{tabular}{l rr rr rr}
\toprule
 & \multicolumn{2}{c}{$K = 100$} & \multicolumn{2}{c}{Mip360} & \multicolumn{2}{c}{ScanNet} \\
\cmidrule(lr){2-3}\cmidrule(lr){4-5}\cmidrule(lr){6-7}
Variant & @30 & Time & @30 & Time & @30 & Time \\
\midrule
DejaView, dense & 82.3 & 2.10 & 97.5 & 14.1 & 86.5 & 319.6 \\
\midrule
top-$256$  & 68.8 & 1.85 & 88.4 & 5.6 & 75.1 & 33.6 \\
$k$-means on raw tokens & 65.3 & 1.39 & 86.7 & 4.4 & 71.9 & 30.3 \\
\midrule
w/o reference view $\mathcal{R}$ & 74.1 & 1.30 & 93.1 & 4.0 & 84.7 & 25.6 \\
w/o summaries $\mathcal{M}(i)$ & 76.3 & 1.29 & 95.0 & 3.9 & 83.5 & 24.3 \\
w/o $\mathcal{R}$ and $\mathcal{M}(i)$ & 65.1 & 1.26 & 86.4 & 3.8 & 79.2 & 23.8 \\
\midrule
\textbf{\methodDV} & \textbf{78.2} & \textbf{1.32} & \textbf{96.0} & \textbf{4.1} & \textbf{84.6} & \textbf{26.1} \\
\bottomrule
\end{tabular}
\end{table}

\subsection{Ablation study}
\label{sec:ablations}

In Tab.~\ref{tab:abl_index}, we provide an ablation study regarding different aspects of our method. First, we investigate \emph{which keys a patch attends to.} Attending to every patch of the nearest clusters is the most accurate choice. Keeping only the $256$ of them with the highest adjacency scores, as a per-query top-$K$ retrieval would, loses significant accuracy of AUC@30 in every setting. It is also slower, since selecting the keys costs more than attending to fewer of them saves. Second, we investigate \emph{what the clusters should be built on.} Running $k$-means on the raw tokens instead of the learned embeddings $e_i$ is the largest single loss. This means that we need a good geometric representation of tokens, and the contrastive projection is what finds them. Third, we investigate \emph{the effectiveness of coarse key sets.} Without the reference view, AUC@30 drops significantly at $K = 100$ views, and it also drops noticeably without the cluster summaries. Without both, it drops far more than the sum, to the level of raw-token clustering. This is because each set gives a patch a path to the scene beyond its nearest clusters, and at least one is needed. Together they add only $P + \sqrt{N}$ keys per patch and only $5$ to $9\%$ of the time.

%% file: sec/5_conclusion.tex
\section{Conclusion and Discussion}
\label{sec:conclusion}
In this paper, we presented \method{}, a general method that replaces the dense global attention of feed-forward reconstruction models with geometry-guided sparse attention. A learned projection embeds each patch such that patches observing the same surface lie close together, and our sparse geometric attention then selects a small subset of relevant keys for each query by clustering in this embedding space. We applied \method{} to two state-of-the-art dense-attention models and showed that it is the fastest model at every sequence length, up to $12\times$ faster than dense attention at $1{,}000$ views, and the most accurate of the efficient alternatives, with accuracy comparable to the original dense models. Despite the good performance, \method{} also comes with limitations: sparsification still brings a small drop in accuracy relative to dense attention. Closing this gap is a promising direction for future work.

%% file: main.bbl
\begin{thebibliography}{75}
\providecommand{\natexlab}[1]{#1}
\providecommand{\url}[1]{\texttt{#1}}
\expandafter\ifx\csname urlstyle\endcsname\relax
  \providecommand{\doi}[1]{doi: #1}\else
  \providecommand{\doi}{doi: \begingroup \urlstyle{rm}\Url}\fi

\bibitem[Arthur and Vassilvitskii(2007)]{Arthur07}
David Arthur and Sergei Vassilvitskii.
\newblock k-means++: The advantages of careful seeding.
\newblock In \emph{ACM-SIAM Symposium on Discrete Algorithms}, 2007.

\bibitem[Avetisyan et~al.(2024)Avetisyan, Xie, Howard-Jenkins, Yang, Aroudj,
  Patra, Zhang, Frost, Holland, Orme, et~al.]{Avetisyan24}
Armen Avetisyan, Christopher Xie, Henry Howard-Jenkins, Tsun-Yi Yang, Samir
  Aroudj, Suvam Patra, Fuyang Zhang, Duncan Frost, Luke Holland, Campbell Orme,
  et~al.
\newblock {SceneScript}: Reconstructing scenes with an autoregressive
  structured language model.
\newblock In \emph{European Conference on Computer Vision}, pages 247--263,
  2024.

\bibitem[Azinovi\'c et~al.(2022)Azinovi\'c, Martin-Brualla, Goldman,
  Nie{\ss}ner, and Thies]{Azinovic22}
Dejan Azinovi\'c, Ricardo Martin-Brualla, Dan~B. Goldman, Matthias Nie{\ss}ner,
  and Justus Thies.
\newblock Neural {RGB-D} surface reconstruction.
\newblock In \emph{Conference on Computer Vision and Pattern Recognition},
  2022.

\bibitem[Barron et~al.(2022)Barron, Mildenhall, Verbin, Srinivasan, and
  Hedman]{Barron22}
J. Barron, B. Mildenhall, D. Verbin, P.~P. Srinivasan, and P. Hedman.
\newblock {Mip-Nerf 360: Unbounded Anti-Aliased Neural Radiance Fields}.
\newblock In \emph{Conference on Computer Vision and Pattern Recognition},
  2022.

\bibitem[Beltagy et~al.(2020)Beltagy, Peters, and Cohan]{Beltagy20}
Iz Beltagy, Matthew~E. Peters, and Arman Cohan.
\newblock Longformer: The long-document transformer.
\newblock \emph{arXiv Preprint}, 2004.05150, 2020.

\bibitem[Besl and Mckay(1992)]{Besl92}
P. Besl and N. Mckay.
\newblock {A Method for Registration of 3D Shapes}.
\newblock \emph{IEEE Transactions on Pattern Analysis and Machine
  Intelligence}, 14\penalty0 (2):\penalty0 239--256, 1992.

\bibitem[Bolya et~al.(2023)Bolya, Fu, Dai, Zhang, Feichtenhofer, and
  Hoffman]{Bolya23}
Daniel Bolya, Cheng-Yang Fu, Xiaoliang Dai, Peizhao Zhang, Christoph
  Feichtenhofer, and Judy Hoffman.
\newblock Token merging: Your {ViT} but faster.
\newblock In \emph{International Conference on Learning Representations}, 2023.

\bibitem[Burzio et~al.(2026)Burzio, Fischer, Elflein, Zhou, de~Lutio, Ren,
  Huang, Huang, Pollefeys, Leal-Taix\'e, Gojcic, and Turki]{Burzio26}
Alessandro Burzio, Tobias Fischer, Sven Elflein, Qunjie Zhou, Riccardo de
  Lutio, Jiawei Ren, Jiahui Huang, Shengyu Huang, Marc Pollefeys, Laura
  Leal-Taix\'e, Zan Gojcic, and Haithem Turki.
\newblock D\'ej\`a view: Looping transformers for multi-view 3d reconstruction.
\newblock \emph{arXiv Preprint}, 2605.30215, 2026.

\bibitem[Cabon et~al.(2020)Cabon, Murray, and Humenberger]{Cabon20}
Yohann Cabon, Naila Murray, and Martin Humenberger.
\newblock Virtual {KITTI} 2.
\newblock \emph{arXiv preprint arXiv:2001.10773}, 2020.

\bibitem[Chen et~al.(2020)Chen, Kornblith, Norouzi, and Hinton]{Chen20a}
T. Chen, S. Kornblith, M. Norouzi, and G. Hinton.
\newblock {A Simple Framework for Contrastive Learning of Visual
  Representations}.
\newblock In \emph{International Conference on Machine Learning}, 2020.

\bibitem[Chen et~al.(2026)Chen, Qiu, Li, Patrikar, and Scherer]{Chen26}
Yutian Chen, Yuheng Qiu, Ruogu Li, Jay Patrikar, and Sebastian Scherer.
\newblock Co-me: Confidence guided token merging for visual geometric
  transformers.
\newblock In \emph{Conference on Computer Vision and Pattern Recognition},
  pages 14590--14599, 2026.

\bibitem[Child et~al.(2019)Child, Gray, Radford, and Sutskever]{Child19}
Rewon Child, Scott Gray, Alec Radford, and Ilya Sutskever.
\newblock Generating long sequences with sparse transformers.
\newblock \emph{arXiv Preprint}, 1904.10509, 2019.

\bibitem[Choy et~al.(2016)Choy, Gwak, Savarese, and Chandraker]{Choy16a}
C. Choy, J. Gwak, S. Savarese, and M. Chandraker.
\newblock {Universal Correspondence Network}.
\newblock In \emph{Advances in Neural Information Processing Systems}, 2016.

\bibitem[Dai et~al.(2017)Dai, Chang, Savva, Halber, Funkhouser, and
  Nie{\ss}ner]{Dai17b}
A. Dai, A.X. Chang, M. Savva, M. Halber, T. Funkhouser, and M. Nie{\ss}ner.
\newblock {Scannet: Richly-Annotated 3D Reconstructions of Indoor Scenes}.
\newblock In \emph{Conference on Computer Vision and Pattern Recognition},
  2017.

\bibitem[Darcet et~al.(2024)Darcet, Oquab, Mairal, and Bojanowski]{Darcet24}
Timoth{\'e}e Darcet, Maxime Oquab, Julien Mairal, and Piotr Bojanowski.
\newblock Vision transformers need registers.
\newblock In \emph{International Conference on Learning Representations}, 2024.

\bibitem[Deng et~al.(2026)Deng, Li, Wang, Wang, Chen, Yang, and Hu]{Deng26}
Jianing Deng, Yuanzhe Li, Jialu Wang, Song Wang, Tianlong Chen, Huanrui Yang,
  and Jingtong Hu.
\newblock Saf3r: Dynamic sparse attention for feed-forward 3d reconstruction
  transformers.
\newblock \emph{arXiv Preprint}, 2607.03612, 2026.

\bibitem[Dhillon and Modha(2001)]{Dhillon01}
I.~S. Dhillon and D.~S. Modha.
\newblock {Concept Decompositions for Large Sparse Text Data Using Clustering}.
\newblock \emph{Machine Learning}, 42\penalty0 (1):\penalty0 143--175, 2001.

\bibitem[Dong et~al.(2025)Dong, Feng, Guessous, Liang, and He]{Dong25}
J. Dong, B. Feng, D. Guessous, Y. Liang, and H. He.
\newblock {Flex Attention: A Programming Model for Generating Optimized
  Attention Kernels}.
\newblock In \emph{Conference on Machine Learning and Systems}, 2025.

\bibitem[Edstedt et~al.(2024)Edstedt, Sun, B{\"o}kman, Wadenb{\"a}ck, and
  Felsberg]{edstedt2024roma}
Johan Edstedt, Qiyu Sun, Georg B{\"o}kman, M{\aa}rten Wadenb{\"a}ck, and
  Michael Felsberg.
\newblock Roma: Robust dense feature matching.
\newblock In \emph{2024 IEEE/CVF Conference on Computer Vision and Pattern
  Recognition (CVPR)}, pages 19790--19800. IEEE, 2024.

\bibitem[Elflein et~al.(2026)Elflein, Li, Agostinho, Gojcic, Leal-Taix{\'e},
  Zhou, and Osep]{Elflein26}
Sven Elflein, Ruilong Li, S{\'e}rgio Agostinho, Zan Gojcic, Laura
  Leal-Taix{\'e}, Qunjie Zhou, and Aljosa Osep.
\newblock Vgg-t\textsuperscript{3}: Offline feed-forward 3d reconstruction at
  scale.
\newblock In \emph{Conference on Computer Vision and Pattern Recognition},
  2026.

\bibitem[Greff et~al.(2022)Greff, Belletti, Beyer, Doersch, Du, Duckworth,
  Fleet, Gnanapragasam, Golemo, Herrmann, et~al.]{Greff22}
Klaus Greff, Francois Belletti, Lucas Beyer, Carl Doersch, Yilun Du, Daniel
  Duckworth, David~J. Fleet, Dan Gnanapragasam, Florian Golemo, Charles
  Herrmann, et~al.
\newblock {Kubric}: A scalable dataset generator.
\newblock In \emph{Conference on Computer Vision and Pattern Recognition},
  pages 3749--3761, 2022.

\bibitem[Huang et~al.(2018)Huang, Matzen, Kopf, Ahuja, and Huang]{Huang18b}
P. Huang, K. Matzen, J. Kopf, N. Ahuja, and J. Huang.
\newblock {Deepmvs: Learning Multi-View Stereopsis}.
\newblock In \emph{Conference on Computer Vision and Pattern Recognition},
  pages 2821--2830, 2018.

\bibitem[Jin et~al.(2026)Jin, Wu, Zhang, Gao, Barron, Snavely, and
  Holynski]{Jin26}
Haian Jin, Rundi Wu, Tianyuan Zhang, Ruiqi Gao, Jonathan~T. Barron, Noah
  Snavely, and Aleksander Holynski.
\newblock {ZipMap}: Linear-time stateful {3D} reconstruction with test-time
  training.
\newblock In \emph{Conference on Computer Vision and Pattern Recognition},
  2026.

\bibitem[Khosla et~al.(2020)Khosla, Teterwak, Wang, Sarna, Tian, Isola,
  Maschinot, Liu, and Krishnan]{Khosla20}
Prannay Khosla, Piotr Teterwak, Chen Wang, Aaron Sarna, Yonglong Tian, Phillip
  Isola, Aaron Maschinot, Ce Liu, and Dilip Krishnan.
\newblock Supervised contrastive learning.
\newblock In \emph{Advances in Neural Information Processing Systems}, 2020.

\bibitem[Kitaev et~al.(2020)Kitaev, Kaiser, and Levskaya]{Kitaev20}
Nikita Kitaev, {\L}ukasz Kaiser, and Anselm Levskaya.
\newblock Reformer: The efficient transformer.
\newblock In \emph{International Conference on Learning Representations}, 2020.

\bibitem[Lazarow et~al.(2025)Lazarow, Griffiths, Kohavi, Crespo, and
  Dehghan]{Lazarow25}
Justin Lazarow, David Griffiths, Gefen Kohavi, Francisco Crespo, and Afshin
  Dehghan.
\newblock {Cubify Anything}: Scaling indoor {3D} object detection.
\newblock In \emph{Conference on Computer Vision and Pattern Recognition},
  pages 22225--22233, 2025.

\bibitem[Leroy et~al.(2024)Leroy, Cabon, and Revaud]{Leroy24}
Vincent Leroy, Yohann Cabon, and J\'{e}r\^{o}me Revaud.
\newblock Grounding image matching in 3d with {MASt3R}.
\newblock In \emph{European Conference on Computer Vision}, 2024.

\bibitem[Li and Snavely(2018)]{Li18o}
Zhengqi Li and Noah Snavely.
\newblock Megadepth: Learning single-view depth prediction from internet
  photos.
\newblock In \emph{Conference on Computer Vision and Pattern Recognition},
  2018.

\bibitem[Lin et~al.(2026)Lin, Chen, Liew, Chen, Li, Zhao, Peng, Guo, Zhou, Shi,
  Feng, and Kang]{Lin26}
Haotong Lin, Sili Chen, Jun~Hao Liew, Donny~Y. Chen, Zhenyu Li, Yang Zhao, Sida
  Peng, Hengkai Guo, Xiaowei Zhou, Guang Shi, Jiashi Feng, and Bingyi Kang.
\newblock Depth anything 3: Recovering the visual space from any views.
\newblock In \emph{International Conference on Learning Representations}, 2026.

\bibitem[Lindenberger et~al.(2023)Lindenberger, Sarlin, and
  Pollefeys]{Lindenberger23}
Philipp Lindenberger, Paul-Edouard Sarlin, and Marc Pollefeys.
\newblock {LightGlue}: Local feature matching at light speed.
\newblock In \emph{International Conference on Computer Vision}, 2023.

\bibitem[Liu et~al.(2021)Liu, Lin, Cao, Hu, Wei, Zhang, Lin, and Guo]{Liu21a}
Ze Liu, Yutong Lin, Yue Cao, Han Hu, Yixuan Wei, Zheng Zhang, Stephen Lin, and
  Baining Guo.
\newblock Swin transformer: Hierarchical vision transformer using shifted
  windows.
\newblock In \emph{International Conference on Computer Vision}, 2021.

\bibitem[{Mapillary}(2021)]{Mapillary21b}
{Mapillary}.
\newblock {Mapillary Metropolis}.
\newblock \url{https://www.mapillary.com/dataset/metropolis}, 2021.

\bibitem[Mehl et~al.(2023)Mehl, Schmalfuss, Jahedi, Nalivayko, and
  Bruhn]{Mehl23}
Lukas Mehl, Jenny Schmalfuss, Azin Jahedi, Yaroslava Nalivayko, and Andr{\'e}s
  Bruhn.
\newblock {Spring}: A high-resolution high-detail dataset and benchmark for
  scene flow, optical flow and stereo.
\newblock In \emph{Conference on Computer Vision and Pattern Recognition},
  pages 4981--4991, 2023.

\bibitem[Moulon et~al.(2016)Moulon, Monasse, Perrot, and Marlet]{Moulon16}
Pierre Moulon, Pascal Monasse, Romuald Perrot, and Renaud Marlet.
\newblock {OpenMVG}: Open multiple view geometry.
\newblock In \emph{International Workshop on Reproducible Research in Pattern
  Recognition}, pages 60--74. Springer, 2016.

\bibitem[Oord et~al.(2018)Oord, Li, and Vinyals]{Oord18}
Aaron van~den Oord, Yazhe Li, and Oriol Vinyals.
\newblock {Representation Learning with Contrastive Predictive Coding}.
\newblock In \emph{arXiv Preprint}, 2018.

\bibitem[{OpenMVS Contributors}(2026)]{OpenMVS26}
{OpenMVS Contributors}.
\newblock {OpenMVS}.
\newblock \url{https://github.com/cdcseacave/openMVS}, 2026.
\newblock Accessed: 2026-05-06.

\bibitem[Pan et~al.(2024)Pan, Bar{\'a}th, Pollefeys, and
  Sch{\"o}nberger]{Pan24a}
L. Pan, D. Bar{\'a}th, M. Pollefeys, and J.~L. Sch{\"o}nberger.
\newblock {Global Structure-From-Motion Revisited}.
\newblock In \emph{European Conference on Computer Vision}, pages 58--77, 2024.

\bibitem[Pix4D()]{Pix4D}
Pix4D.
\newblock {Pix4D}, 2012.
\newblock https://www.pix4d.com/.

\bibitem[Radford et~al.(2021)Radford, Kim, Hallacy, Ramesh, Goh, Agarwal,
  Sastry, Askell, Mishkin, Clark, Krueger, and Sutskever]{Radford21}
Alec Radford, Jong~Wook Kim, Chris Hallacy, Aditya Ramesh, Gabriel Goh,
  Sandhini Agarwal, Girish Sastry, Amanda Askell, Pamela Mishkin, Jack Clark,
  Gretchen Krueger, and Ilya Sutskever.
\newblock Learning transferable visual models from natural language
  supervision.
\newblock In \emph{Proceedings of the 38th International Conference on Machine
  Learning}, pages 8748--8763. PMLR, 2021.

\bibitem[Reizenstein et~al.(2021)Reizenstein, Shapovalov, Henzler, Sbordone,
  Labatut, and Novotny]{Reizenstein21}
Jeremy Reizenstein, Roman Shapovalov, Philipp Henzler, Luca Sbordone, Patrick
  Labatut, and David Novotny.
\newblock Common objects in {3D}: Large-scale learning and evaluation of
  real-life {3D} category reconstruction.
\newblock In \emph{International Conference on Computer Vision}, pages
  10881--10891, 2021.

\bibitem[Ren et~al.(2026)Ren, Tan, and Han]{Ren26}
Weining Ren, Xiao Tan, and Kai Han.
\newblock Speed3r: Sparse feed-forward 3d reconstruction models.
\newblock In \emph{Proceedings of the IEEE/CVF Conference on Computer Vision
  and Pattern Recognition (CVPR) Findings}, 2026.

\bibitem[Roberts et~al.(2021)Roberts, Ramapuram, Ranjan, Kumar, Bautista,
  Paczan, Webb, and Susskind]{Roberts21}
Mike Roberts, Jason Ramapuram, Anurag Ranjan, Atulit Kumar, Miguel~Angel
  Bautista, Nathan Paczan, Russ Webb, and Joshua~M. Susskind.
\newblock {Hypersim}: A photorealistic synthetic dataset for holistic indoor
  scene understanding.
\newblock In \emph{International Conference on Computer Vision}, 2021.

\bibitem[Roy et~al.(2021)Roy, Saffar, Vaswani, and Grangier]{Roy21}
Aurko Roy, Mohammad Saffar, Ashish Vaswani, and David Grangier.
\newblock Efficient content-based sparse attention with routing transformers.
\newblock \emph{Transactions of the Association for Computational Linguistics},
  9:\penalty0 53--68, 2021.

\bibitem[Sarlin et~al.(2020)Sarlin, DeTone, Malisiewicz, and
  Rabinovich]{Sarlin20}
P.E. Sarlin, D. DeTone, T. Malisiewicz, and A. Rabinovich.
\newblock {Superglue: Learning Feature Matching with Graph Neural Networks}.
\newblock In \emph{Conference on Computer Vision and Pattern Recognition},
  2020.

\bibitem[Sarlin et~al.(2022)Sarlin, Dusmanu, Sch{\"o}nberger, Speciale, Gruber,
  Larsson, Miksik, and Pollefeys]{Sarlin22}
P.-E. Sarlin, M. Dusmanu, J.~L. Sch{\"o}nberger, P. Speciale, L. Gruber, V.
  Larsson, O. Miksik, and M. Pollefeys.
\newblock {LaMAR: Benchmarking Localization and Mapping for Augmented Reality}.
\newblock In \emph{European Conference on Computer Vision}, 2022.

\bibitem[Sch{\"o}nberger and Frahm(2016)]{Schoenberger16a}
J.L. Sch{\"o}nberger and J.M. Frahm.
\newblock {Structure-From-Motion Revisited}.
\newblock In \emph{Conference on Computer Vision and Pattern Recognition},
  2016.

\bibitem[Sch{\"o}ps et~al.(2017)Sch{\"o}ps, Sch{\"o}nberger, Gallian, Sattler,
  Schindler, Pollefeys, and Geiger]{Schops17}
T. Sch{\"o}ps, J. Sch{\"o}nberger, S. Gallian, T. Sattler, K. Schindler, M.
  Pollefeys, and A. Geiger.
\newblock {A Multi-View Stereo Benchmark with High-Resolution Images and
  Multi-Camera Videos}.
\newblock In \emph{Conference on Computer Vision and Pattern Recognition},
  2017.

\bibitem[Shen et~al.(2026)Shen, Zhang, Qu, Zheng, Ji, Zhang, and Cao]{Shen26}
You Shen, Zhipeng Zhang, Yansong Qu, Xiawu Zheng, Jiayi Ji, Shengchuan Zhang,
  and Liujuan Cao.
\newblock Fast{VGGT}: Fast visual geometry transformer.
\newblock In \emph{International Conference on Learning Representations}, 2026.

\bibitem[Shotton et~al.(2013)Shotton, Glocker, Zach, Izadi, Criminisi, and
  Fitzgibbon]{Shotton13c}
J. Shotton, B. Glocker, C. Zach, S. Izadi, A. Criminisi, and A. Fitzgibbon.
\newblock {Scene Coordinate Regression Forests for Camera Relocalization in
  RGB-D Images}.
\newblock In \emph{Conference on Computer Vision and Pattern Recognition},
  pages 2930--2937, 2013.

\bibitem[Straub et~al.(2019)Straub, Whelan, Ma, Chen, Wijmans, Green, Engel,
  Mur-Artal, Ren, Verma, et~al.]{Straub19}
Julian Straub, Thomas Whelan, Lingni Ma, Yufan Chen, Erik Wijmans, Simon Green,
  Jakob~J. Engel, Raul Mur-Artal, Carl Ren, Shobhit Verma, et~al.
\newblock The {Replica} dataset: A digital replica of indoor spaces.
\newblock \emph{arXiv preprint arXiv:1906.05797}, 2019.

\bibitem[Sun et~al.(2021)Sun, Shen, Wang, Bao, and Zhou]{Sun21a}
J. Sun, Z. Shen, Y. Wang, H. Bao, and X. Zhou.
\newblock {LoFTR: Detector-Free Local Feature Matching with Transformers}.
\newblock In \emph{Conference on Computer Vision and Pattern Recognition},
  2021.

\bibitem[Sun et~al.(2026)Sun, Zhu, Lou, Yang, Tang, Zhang, Wang, and
  Zhang]{Sun26}
Xianbing Sun, Zhikai Zhu, Zhengyu Lou, Bo Yang, Jinyang Tang, Liqing Zhang, He
  Wang, and Jianfu Zhang.
\newblock Avggt: Rethinking global attention for accelerating vggt.
\newblock In \emph{Conference on Computer Vision and Pattern Recognition},
  pages 251--260, 2026.

\bibitem[Tola et~al.(2012)Tola, Strecha, and Fua]{Tola12}
E. Tola, C. Strecha, and P. Fua.
\newblock {Efficient Large Scale Multi-View Stereo for Ultra High Resolution
  Image Sets}.
\newblock \emph{Machine Vision and Applications}, 23\penalty0 (5), 2012.

\bibitem[Tosi et~al.(2021)Tosi, Liao, Schmitt, and Geiger]{Tosi21}
Fabio Tosi, Yiyi Liao, Carolin Schmitt, and Andreas Geiger.
\newblock {SMD-Nets}: Stereo mixture density networks.
\newblock In \emph{Conference on Computer Vision and Pattern Recognition},
  pages 8938--8948, 2021.

\bibitem[Tyszkiewicz et~al.(2020)Tyszkiewicz, Fua, and
  Trulls]{tyszkiewicz2020disk}
Micha{\l} Tyszkiewicz, Pascal Fua, and Eduard Trulls.
\newblock Disk: Learning local features with policy gradient.
\newblock \emph{Advances in neural information processing systems},
  33:\penalty0 14254--14265, 2020.

\bibitem[Umeyama(1991)]{Umeyama91}
S. Umeyama.
\newblock {Least-Squares Estimation of Transformation Parameters Between Two
  Point Patterns}.
\newblock \emph{IEEE Transactions on Pattern Analysis and Machine
  Intelligence}, 13\penalty0 (4), 1991.

\bibitem[Van~Hoorick et~al.(2024)Van~Hoorick, Wu, Ozguroglu, Sargent, Liu,
  Tokmakov, Dave, Zheng, and Vondrick]{VanHoorick24}
Basile Van~Hoorick, Rundi Wu, Ege Ozguroglu, Kyle Sargent, Ruoshi Liu, Pavel
  Tokmakov, Achal Dave, Changxi Zheng, and Carl Vondrick.
\newblock Generative camera dolly: Extreme monocular dynamic novel view
  synthesis.
\newblock In \emph{European Conference on Computer Vision}, pages 313--331,
  2024.

\bibitem[Wang et~al.(2026{\natexlab{a}})Wang, Schmidt, Piekenbrinck, and
  Leibe]{Wang26d}
Chung-Shien~Brian Wang, Christian Schmidt, Jens Piekenbrinck, and Bastian
  Leibe.
\newblock Block-sparse global attention for efficient multi-view geometry
  transformers.
\newblock In \emph{Conference on Computer Vision and Pattern Recognition},
  pages 14546--14555, 2026{\natexlab{a}}.

\bibitem[Wang and Agapito(2025)]{Wang25d}
Hengyi Wang and Lourdes Agapito.
\newblock {3D} reconstruction with spatial memory.
\newblock In \emph{International Conference on 3D Vision}, 2025.

\bibitem[Wang et~al.(2025{\natexlab{a}})Wang, Chen, Karaev, Vedaldi, Rupprecht,
  and Novotny]{Wang25}
Jianyuan Wang, Minghao Chen, Nikita Karaev, Andrea Vedaldi, Christian
  Rupprecht, and David Novotny.
\newblock Vggt: Visual geometry grounded transformer.
\newblock In \emph{Conference on Computer Vision and Pattern Recognition},
  2025{\natexlab{a}}.

\bibitem[Wang et~al.(2026{\natexlab{b}})Wang, Chen, Zhang, Karaev,
  Sch{\"o}nberger, Labatut, Bojanowski, Novotny, Vedaldi, and
  Rupprecht]{Wang26b}
Jianyuan Wang, Minghao Chen, Shangzhan Zhang, Nikita Karaev, Johannes
  Sch{\"o}nberger, Patrick Labatut, Piotr Bojanowski, David Novotny, Andrea
  Vedaldi, and Christian Rupprecht.
\newblock {VGGT-$\Omega$}.
\newblock In \emph{Conference on Computer Vision and Pattern Recognition},
  2026{\natexlab{b}}.

\bibitem[Wang et~al.(2025{\natexlab{b}})Wang, Zhang, Holynski, Efros, and
  Kanazawa]{Wang25c}
Qianqian Wang, Yifei Zhang, Aleksander Holynski, Alexei~A. Efros, and Angjoo
  Kanazawa.
\newblock Continuous {3D} perception model with persistent state.
\newblock In \emph{Conference on Computer Vision and Pattern Recognition},
  2025{\natexlab{b}}.

\bibitem[Wang et~al.(2024)Wang, Leroy, Cabon, Chidlovskii, and Revaud]{Wang24}
Shuzhe Wang, Vincent Leroy, Yohann Cabon, Boris Chidlovskii, and Jerome Revaud.
\newblock Dust3r: Geometric 3d vision made easy.
\newblock \emph{arXiv Preprint}, 2312.14132, 2024.

\bibitem[Wang et~al.(2020)Wang, Zhu, Wang, Hu, Qiu, Wang, Hu, Kapoor, and
  Scherer]{Wang20n}
Wenshan Wang, Delong Zhu, Xiangwei Wang, Yaoyu Hu, Yuheng Qiu, Chen Wang, Yafei
  Hu, Ashish Kapoor, and Sebastian Scherer.
\newblock {TartanAir}: A dataset to push the limits of visual {SLAM}.
\newblock In \emph{International Conference on Intelligent Robots and Systems},
  2020.

\bibitem[Wang et~al.(2025{\natexlab{c}})Wang, Zhou, Zhu, et~al.]{Wang25b}
Yifan Wang, Jianjun Zhou, Haoyi Zhu, et~al.
\newblock $\pi^3$: Scalable permutation-equivariant visual geometry learning.
\newblock \emph{arXiv Preprint}, 2507.13347, 2025{\natexlab{c}}.

\bibitem[Wang and Xu(2026)]{Wang26c}
Zipeng Wang and Dan Xu.
\newblock Flashvggt: Efficient and scalable visual geometry transformers with
  compressed descriptor attention.
\newblock In \emph{Conference on Computer Vision and Pattern Recognition},
  pages 21826--21835, 2026.

\bibitem[Weinberger and Saul(2008)]{Weinberger08}
K.Q. Weinberger and L.K. Saul.
\newblock {Fast Solvers and Efficient Implementations for Distance Metric
  Learning}.
\newblock In \emph{International Conference on Machine Learning}, 2008.

\bibitem[Wu et~al.(2022)Wu, Rabe, Hutchins, and Szegedy]{Wu22}
Yuhuai Wu, Markus~N. Rabe, DeLesley Hutchins, and Christian Szegedy.
\newblock Memorizing transformers.
\newblock In \emph{International Conference on Learning Representations}, 2022.

\bibitem[Xiang et~al.(2025)Xiang, Lv, Xu, Deng, Wang, Zhang, Chen, Tong, and
  Yang]{Xiang25a}
Jianfeng Xiang, Zelong Lv, Sicheng Xu, Yu Deng, Ruicheng Wang, Bowen Zhang,
  Dong Chen, Xin Tong, and Jiaolong Yang.
\newblock Structured 3d latents for scalable and versatile 3d generation.
\newblock In \emph{Conference on Computer Vision and Pattern Recognition},
  2025.

\bibitem[Yang et~al.(2025)Yang, Sax, Liang, Henaff, Tang, Cao, Chai, Meier, and
  Feiszli]{Yang25}
Jianing Yang, Alexander Sax, Kevin~J. Liang, Mikael Henaff, Hao Tang, Ang Cao,
  Joyce Chai, Franziska Meier, and Matt Feiszli.
\newblock {Fast3R}: Towards {3D} reconstruction of 1000+ images in one forward
  pass.
\newblock In \emph{Conference on Computer Vision and Pattern Recognition},
  2025.

\bibitem[Yao et~al.(2020)Yao, Luo, Li, Zhang, Ren, Zhou, Fang, and
  Quan]{Yao20b}
Yao Yao, Zixin Luo, Shiwei Li, Jingyang Zhang, Yufan Ren, Lei Zhou, Tian Fang,
  and Long Quan.
\newblock {BlendedMVS: A Large-Scale Dataset for Generalized Multi-View Stereo
  Networks}.
\newblock In \emph{Conference on Computer Vision and Pattern Recognition},
  2020.

\bibitem[Zaheer et~al.(2020)Zaheer, Guruganesh, Dubey, Ainslie, Alberti,
  Ontanon, Pham, Ravula, Wang, Yang, and Ahmed]{Zaheer20}
Manzil Zaheer, Guru Guruganesh, Kumar~Avinava Dubey, Joshua Ainslie, Chris
  Alberti, Santiago Ontanon, Philip Pham, Anirudh Ravula, Qifan Wang, Li Yang,
  and Amr Ahmed.
\newblock Big bird: Transformers for longer sequences.
\newblock In \emph{Advances in Neural Information Processing Systems}, 2020.

\bibitem[Zamir et~al.(2018)Zamir, Sax, Shen, Guibas, Malik, and
  Savarese]{Zamir18}
Amir~R. Zamir, Alexander Sax, William Shen, Leonidas~J. Guibas, Jitendra Malik,
  and Silvio Savarese.
\newblock {Taskonomy}: Disentangling task transfer learning.
\newblock In \emph{Conference on Computer Vision and Pattern Recognition},
  pages 3712--3722, 2018.

\bibitem[Zhang et~al.(2026)Zhang, Zhang, Zhou, Zhao, and Tang]{Zhang26}
Haoyu Zhang, Zeyu Zhang, Zedong Zhou, Yang Zhao, and Hao Tang.
\newblock Lite3r: A model-agnostic framework for efficient feed-forward 3d
  reconstruction.
\newblock \emph{arXiv Preprint}, 2605.11354, 2026.

\bibitem[Zhuo et~al.(2025)Zhuo, Zheng, Guo, Wu, Zhou, and Lu]{Zhuo25}
Dong Zhuo, Wenzhao Zheng, Jiahe Guo, Yuqi Wu, Jie Zhou, and Jiwen Lu.
\newblock Streaming 4d visual geometry transformer.
\newblock \emph{arXiv Preprint}, 2507.11539, 2025.

\end{thebibliography}
